\pdfoutput=1

\documentclass[11pt]{article}

\usepackage{ACL2023}

\usepackage{times}
\usepackage{latexsym}

\usepackage[T1]{fontenc}

\usepackage[utf8]{inputenc}

\usepackage{microtype}

\usepackage{inconsolata}

\title{The Impact of Data Corruption on Named Entity Recognition for Low-resourced Languages}

\author{\textbf{Manuel Fokam}  \and \textbf{Michael Beukman} \\
        School of Computer Science and Applied Mathematics, \\ 
        University of the Witwatersrand, Johannesburg, South Africa \\ 
        \{arnolfokam23,mcbeukman\}@gmail.com}

\usepackage{graphicx}
\usepackage{float}
\usepackage{subcaption}
\usepackage{adjustbox}
\usepackage{booktabs}
\usepackage{multirow}
\usepackage{comment}
\usepackage{amsmath}
\usepackage{amsthm}
\usepackage[english]{babel}
\usepackage{xspace}
\usepackage{latexsym}
\usepackage{xcolor}

\newtheorem{exahttps://www.overleaf.com/project/63904fa168a9cddc37b275dd/detachermple}{Example}
\makeatletter
\newcommand\footnoteref[1]{\protected@xdef\@thefnmark{\ref{#1}}\@footnotemark}
\makeatother

\newcommand{\afriberta}{AfriBERTa\xspace}
\newcommand{\afroxlmr}{Afro-XLM-R\xspace}
\newcommand{\xlmr}{XLM-R\xspace}
\newcommand{\myO}{\textit{O}\xspace}
\addto\extrasenglish{  
    
}
\addto\extrasenglish{  
    
}
\addto\extrasenglish{  
    
}

\begin{document}

\maketitle

\begin{abstract}
    Data availability and quality are major challenges in natural language processing for low-resourced languages.
    In particular, there is significantly less data available than for higher-resourced languages. This data is also often of low quality, rife with errors, invalid text or incorrect annotations. Many prior works focus on dealing with these problems, either by generating synthetic data, or filtering out low-quality parts of datasets. We instead investigate these factors more deeply, by systematically measuring the effect of data quantity and quality on the performance of pre-trained language models in a low-resourced setting. Our results show that having fewer completely-labelled sentences is significantly better than having more sentences with missing labels; and that models can perform remarkably well with only 10\% of the training data. Importantly, these results are consistent across ten low-resource languages, English, and four pre-trained models.
    
\end{abstract}

\section{Introduction} 
Natural Language Processing (NLP) is a rapidly growing field that has been applied to a wide range of tasks and domains~\citep{vaswani2017Attention,conneau2019Unsupervised}. However, much of the focus in NLP has been on high-resource languages such as English~\citep{vaswani2017Attention,radford2018Improving,radford2019language_gpt2}. While this has led to notable advancements for these languages, low-resourced languages have not received as much attention, resulting in a significant performance gap between high- and low-resourced languages. This has prompted an increasing number of studies focused exclusively on low-resourced languages, resulting in the development of models~\citep{ogueji2021Small,alabi2022Multilingual} and the introduction of datasets~\citep{adelani2021MasakhaNER,adelani2022Thousand,adelani2022Masakhaner}. 

Despite this impressive progress, data remains a limiting factor for low-resourced NLP~\citep{adelani2022Thousand,adelani2022Masakhaner}. In particular, the two main problems are the availability and quality of data. First, the datasets available for low-resourced languages tend to be smaller than those for high-resourced languages, and for many languages, no data exists at all~\citep{martinus2019Focus,adelani2021MasakhaNER}. Secondly, the available datasets are often of questionable quality, containing invalid text or incorrect annotations~\citep{kreutzer2021Quality}, which has detrimental effects on the models trained on these datasets~\citep{abdul2012Extrinsic,alabi2019Massive}.

This means that many existing datasets in low-resourced NLP are either small or of low quality. This observation has led to research that investigates the tradeoff between the amount and quality of data~\citep{gasco-etal-2012-data,alabi2019Massive,de-gibert-bonet-etal-2022-quality}. This line of work has provided valuable insights that allow NLP practitioners to make informed decisions when faced with a choice of which dataset should be used to train a model. However, many of these works focus on comparing different datasets, often from different domains, without clearly quantifiable tradeoffs between data quantity and quality~\citep{alabi2019Massive}. While these approaches can be useful, a more comprehensive and precise approach is needed to fully understand the tradeoff between data quantity and quality in low-resourced NLP. 

To address this, we take a different perspective and focus on systematically and quantifiably reducing the quality of datasets and examining the effects of this on the performance of NLP models. Additionally, by altering the amount of data used to train our models, we can compare the tradeoffs between quality and quantity. We do this by devising various controllable corruption strategies, and training models on different levels of corrupted data. 
Our focus is on a named-entity recognition task due to its prevalence in many NLP systems and the availability of a few high-quality datasets in low-resourced languages. We fine-tune existing pre-trained language models, as this is a common and high-performing approach, especially for low-resourced languages~\citep{ogueji2021Small,adelani2021MasakhaNER,alabi2022Multilingual}. 

We provide systematic evidence to support prior findings that the quality of data, in general, is strongly preferred over quantity.  Furthermore, our findings are consistent across eleven different languages and four pre-trained models, suggesting that our conclusions hold true in a general sense.
\section{Background and Related Work}
\subsection{Named Entity Recognition}
Named entity recognition (NER) is a token classification task, where the goal is to classify each token or word in a text as an Organisation, Location, Person, Date, or indicate that the token does not correspond to a named entity by giving it the label of ``Other''. NER as a field has many impactful applications in NLP pipelines and use-cases\citep{sang2003introduction_conll,lample2016Playing,adelani2021MasakhaNER}.
A typical NER dataset consists of multiple sentences, with each sentence containing both the words and their associated labels.

The prevailing approach to train NER models is to use a pre-trained large language model (such as BERT~\citep{devlin2019BERT}, XLM-Roberta~\citep{conneau2019Unsupervised}, etc.) and fine-tune it on a small amount of NER data~\citep{conneau2019Unsupervised,adelani2021MasakhaNER}. These models were pre-trained on a large corpus of unlabelled text, and resulted in improved downstream performance after fine-tuning compared to training on NER data from scratch.
The overall classification F1 score, calculated as the harmonic mean of precision and recall, is generally used as the main metric of performance in NER~\citep{sang2003introduction_conll,adelani2021MasakhaNER}.

\subsection{Data Collection and Annotation}
Since the lack of data has traditionally been a major limiting factor for low-resourced NLP research, multiple different approaches have developed to effectively collect data in resource-constrained settings. In particular, community involvement has played a large part in this~\citep{nekoto2020Participatory,nekoto2022Participatory}, where native speakers annotate or create datasets to be used in research. This has led to the creation of many different datasets~\citep{adelani2021MasakhaNER,nekoto2022Participatory}, but it relies on community members instead of trained annotators, which may result in some aspects of the annotation being less accurate. Furthermore, while this approach can successfully develop datasets for low-resourced languages, due to logistic challenges and a limited amount of unlabelled text, these datasets are often significantly smaller than high-resourced datasets~\citep{conneau2019Unsupervised,adelani2021MasakhaNER}.

\subsection{Analysis of Quality vs Quantity in Low-resourced Languages}
While there has been significant progress in recent years, datasets for low-resourced language are often quite small and limited, or exhibit low quality. Both of these factors can lead to poorly-performing models. 
For instance, \citet{kreutzer2021Quality} perform a large-scale audit of several web-scale and automatically extracted multilingual datasets, and find that the quality is often poor, with non-linguistic or otherwise invalid text being commonplace. 

This lack of quality can have great effects on the performance of models. \citet{alabi2019Massive} show that for certain low-resourced African languages, using a significantly smaller, but curated dataset outperforms training a model on a large, but noisy dataset. \citet{abdulmumin2022Separating} find similar results, where training on filtered data of higher quality improved the performance of translation models for low-resourced languages.
Many of these works consider one or two completely different datasets, and compare the relative quality and quantity. This, however, lacks a systematic approach that controls for other factors such as the domain of the data. In addition, the filtering-based approaches often use a learned model as a filter and select only sentences that have a predicted quality value above a certain threshold~\citep{abdulmumin2022Separating,de-gibert-bonet-etal-2022-quality}. While this does provide a quantifiable level of quality, it may not be comparable across datasets or different filtering models. Additionally, datasets may exhibit different levels of certain problems, e.g. some datasets may have many tokens corresponding to punctuation whereas others may have sentences in a different language to the rest of the data. These problems may make it hard to accurately compare the results of these studies and use their conclusions in practice.





\section{Methodology}
\label{sec:method}
Our aim is to analyse and quantify the impact of data corruption on the performance of pre-trained language models. This understanding would enable us to make more informed decisions about the relative importance of data quality and quantity, ultimately leading to improved data creation processes and selection of NLP training data for practitioners.

While we can corrupt NER datasets in various ways, we choose corruptions that simulate a mislabelling scenario during the annotation process, e.g. mislabelling a person in a sentence as an organisation. There are two main reasons for this choice. Firstly, many NER datasets are formed by taking an existing text source, which is usually of high quality, such as news data~\citep{adelani2021MasakhaNER} and annotating each word; thus, errors are more likely to appear during the annotation process. Secondly, it is challenging to corrupt the base sentences in a reasonable, quantifiable and incremental way, as sentences encompass meaning which is often hard to change atomically.

Thus, we focus on corrupting only the labels, using different strategies detailed in \autoref{sec:corrupt_strats}. For each corruption strategy, we uniformly vary the amount of corruption and train our models using the new, corrupted dataset. This process allows us to evaluate how each corruption strategy affects the model as we adjust the degree of corruption. As an additional experiment, we vary the size of the data available to the model by using only a subset of the sentences without corrupting any labels, allowing us to determine the effect of varying the amount of data on performance.

We discuss the data used in this study in \autoref{sec:data}.
 

\begin{figure*}
    \centering
    \includegraphics[width=1\linewidth]{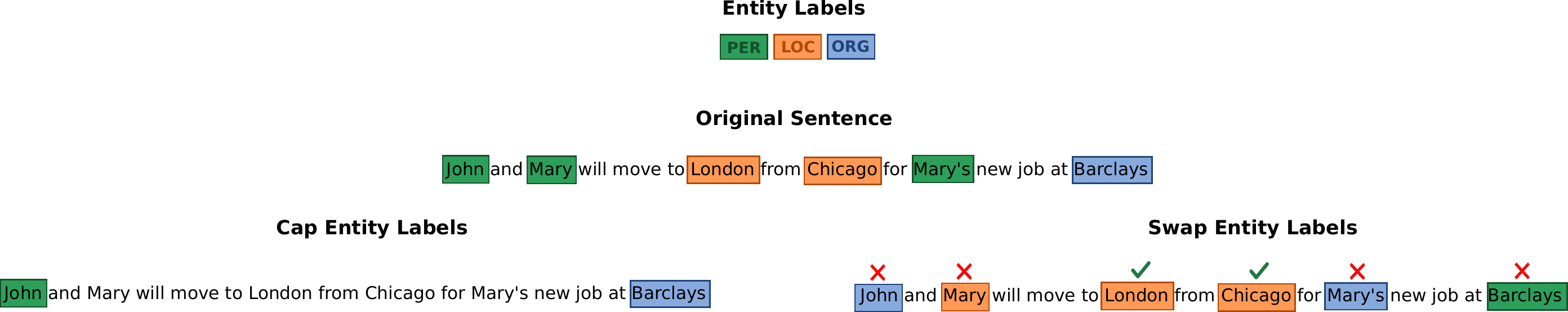}
    \caption{An illustration of the different corruption strategies we use. When (Left) capping labels, we effectively remove a certain fraction of labels, replacing them with \myO. When (Right) swapping labels, we instead randomly replace a label with an incorrect one. This figure is just illustrative, and in our experiments we have a percentage threshold; for instance, corrupting 30\% of the labels across the entire dataset.
    }
    \label{fig:main_fig}
\end{figure*}

\subsection{Different Corruption Strategies}
\label{sec:corrupt_strats}
This section contains descriptions of the corruption strategies that we use, with \autoref{fig:main_fig} providing a visual illustration of our quality-related corruption strategies.
We only change the training data while leaving the evaluation data unchanged to obtain an objective comparison of different corruption strategies.
\subsubsection{Sentence Capping}
\label{sec:method:pruning}

Dataset annotation is generally expensive and logistically challenging when multiple participants are involved. As a result, low-resourced NLP datasets are often not particularly large~\citep{adelani2021MasakhaNER}. Due to this observation, we first evaluate the effect of varying the size of the data available to our NER models. In this strategy, we randomly remove sentences from the original dataset to create sub-datasets with fewer sentences than the original dataset. This process allows us to measure the model performance as a function of data quantity. We choose to represent quantity as a function of the number of sentences because removing words can alter the meaning of a sentence in ways we cannot control.

\subsubsection{Entity Label Capping}
\label{sec:method:cap}

A rich NER dataset would be a dataset with a high annotation density, i.e., a high number of annotated entities per sentence.
This strategy aims at inhibiting the model by thresholding the number of entity annotations allowed in the dataset. In the real world, this would be equivalent to a situation where an annotator failed to label a particular token or span of tokens as one of the entities PER, LOC, ORG, DATE, instead giving it the default entity type \myO, which generally means \textit{not relevant}.

Here we globally corrupt the data by choosing a certain percentage of labels to keep across the entire dataset. For example, $50\%$ would mean that we randomly remove half of all entity labels (replacing them with \myO), which may leave some sentences unmodified and others entirely without annotations.

For this and the following corruption strategy, we consider the atomic element to be a single named entity, even if this consists of multiple words. As a result, we change the entire span of an entity label instead of just a part thereof. 

\subsubsection{Entity Label Swapping}
\label{sec:method:swapping}
Another scenario that could happen during the annotation procedure would be the mislabelling of a span of tokens with the wrong entity. 
For example, A person named \textit{Christian Dior} mistakenly labelled as an organisation due to some bias in the knowledge of the annotator. These mistakes may create datasets with contradictory labels, with the same tokens being used in very similar contexts but labelled differently. Therefore, our goal behind this corruption strategy is to determine how robust large pre-trained language models are to such mistakes. 
Here we again choose a global percentage, randomly selecting labelled entities according to this percentage and swapping their labels with incorrect ones.


\subsection{Data}
\label{sec:data}
We use the MasakhaNER dataset~\citep{adelani2021MasakhaNER}, a high-quality NER dataset for ten low-resourced African languages licensed under CC-BY-4.0-NC. We specifically focus on low-resourced languages, as these languages often suffer from the aforementioned problems. Furthermore, this dataset is of high quality, which allows us to evaluate the full spectrum of quality, from gold-standard to completely corrupted.  

As a baseline, we also use the CONLL NER dataset, which is a staple NER dataset in English~\citep{sang2003introduction_conll}, with many more sentences than any of the MasakhaNER languages. \autoref{tab:data_info} contains information about the number of sentences and entities for each NER corpus.

\begin{table}[h]
    \centering
    \caption{Information about the data. Entity Density refers to the fraction of tokens that are entities. All languages use the Latin script, except Amharic, which uses the Fidel script.}
    \label{tab:data_info}
    \begin{adjustbox}{width=1\linewidth}
    \begin{tabular}{llrrrrr}
\toprule
\multirow{2}{*}{Code} & \multirow{2}{*}{Code} &   \multirow{2}{*}{\# Sentences} &  \multirow{2}{*}{\# Tokens} &  \multirow{2}{*}{\# Entities} &  \multirow{1}{*}{Entities per} &  \% Entities \\
&&&&&Sentence $\text{ }$ & in Tokens
\\ 
\midrule
Amharic         &  amh &         1,750 &     25,829 &        3,995 &                    2.3 &                  15.5 \\
Luganda         &  lug &         1,428 &     33,003 &        5,039 &                    3.5 &                  15.3 \\
Luo             &  luo &           644 &     18,577 &        2,704 &                    4.2 &                  14.6 \\
English         &   en &        14,042 &    203,621 &       29,450 &                    2.1 &                  14.5 \\
Nigerian Pidgin &  pcm &         2,124 &     52,604 &        7,392 &                    3.5 &                  14.1 \\
Kinyarwanda     &  kin &         2,116 &     47,912 &        6,104 &                    2.9 &                  12.7 \\
Swahili         &  swa &         2,109 &     56,599 &        7,161 &                    3.4 &                  12.7 \\
Hausa           &  hau &         1,912 &     55,010 &        6,836 &                    3.6 &                  12.4 \\
Igbo            &  ibo &         2,235 &     42,719 &        5,294 &                    2.4 &                  12.4 \\
Yorùbá          &  yor &         2,171 &     56,274 &        6,324 &                    2.9 &                  11.2 \\
Wolof           &  wol &         1,871 &     36,805 &        2,157 &                    1.2 &                   5.9 \\
\bottomrule
\end{tabular}

    \end{adjustbox}
\end{table}

\begin{table*}
    \centering
    \caption{Information about the different pre-trained language models we use. In the \textit{MasakhaNER Languages} column, we list only the languages the model pre-trained on that are included in the MasakhaNER dataset.}
    \label{tab:plms}
    \begin{adjustbox}{width=1\linewidth}
    \begin{tabular}{llllp{0.4\linewidth}}
        \toprule
        Name & Model Version & Source & Parameters & MasakhaNER Languages \\
        \midrule
        \afriberta & {\texttt{afriberta-large}} & \citet{ogueji2021Small} & 126M  & amh, hau, ibo, kin, pcm, swa, yor \\
        \afroxlmr & \texttt{afro-xlmr-base} & \citet{alabi2022Multilingual} & 270M  & amh, hau, ibo, kin, pcm, swa, yor \\
        XLM Roberta & \texttt{xlm-roberta-base} & \citet{conneau2019Unsupervised} & 270M  & amh, hau, swa \\
        Multilingual BERT & \texttt{bert-base-multilingual-cased} & \citet{devlin2019BERT} & 110M  & swa, yor \\
        \bottomrule
    \end{tabular}
    \end{adjustbox}
\end{table*}

\section{Experiments}
Having described our corruption strategies, we now perform our experiments and showcase our results. We consider the three corruption strategies described above, and use four different pre-trained language models, described in \autoref{sec:diff_models}. Each run consists of fine-tuning a single pre-trained model on a single language's dataset, either the original one or a corrupted version. We fine-tune models for 25 epochs, as the results were similar to 50 epochs (which \citet{adelani2021MasakhaNER} used) and trained much faster. We use a learning rate of $5e-5$, a batch size of $64$, and a sequence length of $200$. We run all experiments over three seeds and average the results. The compute nodes used to run our experiments are equipped with NVIDIA RTX 3090 GPUs.


We specifically investigate the effect of progressively corrupting data on the performance of each model, measured by the overall F1 score. 
This simulates the effect of having low data quality (for instance, due to incorrect annotations), but allows us to study this in a controlled setting. We do not modify the test datasets at all.

Then, to normalise results across models and languages, we divide each F1 score by the value obtained when training the same model on the full, uncorrupted dataset. This effectively measures what fraction of performance is lost when corrupting data and allows us to transform all of the metrics to fall between 0 and 1, resulting in the metrics being comparable across languages and models. 

In the \textit{cap sentences} strategy, where we train models on a subset of data, we specifically remove a certain percentage of the data and train the model on the remaining sentences. Since the specific fraction we keep may have an effect, we run this experiment three times, each time with different random selections of data. We average over these permutations and find that the results are very similar across them.


\subsection{Different Pre-trained Language Models}
\label{sec:diff_models}

We use four different pre-trained language models. We first consider two models developed specifically for low-resourced African languages, \afriberta and \afroxlmr. 
The other two models are traditional multilingual models, with the majority of the training datasets consisting of high-resourced languages, \xlmr and mBERT (multilingual BERT).
\afriberta~\citep{ogueji2021Small} was pre-trained on less than 1GB of African language text. \afroxlmr~\citep{alabi2022Multilingual} used \textit{language adaptive fine-tuning}, where a pre-trained language model is fine-tuned on unlabelled data using the same objective that was used during pre-training. \afroxlmr performed this process on 20 languages, 17 of them from Africa, starting from XLM Roberta. XLM Roberta~\citep{conneau2019Unsupervised} is a high-performing model that was pre-trained on 100 languages. Finally, mBERT~\citep{devlin2019BERT} used the standard BERT training process on 104 languages, using data from Wikipedia.

We choose the specific model versions to be roughly comparable in terms of the number of parameters. More information about the models is included in \autoref{tab:plms}.

\subsection{Initial Results}
In \autoref{tab:performance_all}, we show the results when each model trains on the entire training dataset. Overall, most models perform well on most languages, with \afroxlmr performing the best on average. mBERT, on the other hand, performs the worst overall, with an F1 score of 0 on Amharic, as it was not pre-trained on data containing this script~\citep{adelani2021MasakhaNER}.

\begin{table}
    \centering
    \caption{The F1 score when fine-tuning each model on unaltered training data. Bold indicates the best performance per language.}
    \label{tab:performance_all}
    \begin{adjustbox}{width=1\linewidth}
    \begin{tabular}{lllllr}
\toprule
Model &            AfriBERTa &           Afro-XLM-R &       XLM-R &                mBERT & Average \\
Language &                      &                      &             &                      &         \\
\midrule
amh      &           72.1 (0.9) &  \textbf{75.9 (1.9)} &  71.9 (0.9) &            0.0 (0.0) &    55.0 \\
en       &           88.5 (0.3) &  \textbf{92.8 (0.1)} &  92.7 (0.2) &           92.6 (0.2) &    91.7 \\
hau      &           90.0 (0.5) &  \textbf{90.8 (0.4)} &  89.8 (0.2) &           87.2 (0.5) &    89.5 \\
ibo      &  \textbf{87.1 (0.3)} &           87.0 (0.6) &  83.2 (0.2) &           84.7 (0.5) &    85.5 \\
kin      &           74.1 (0.7) &  \textbf{78.1 (0.2)} &  72.5 (1.3) &           70.7 (0.5) &    73.8 \\
lug      &           78.7 (0.2) &  \textbf{81.3 (0.2)} &  77.7 (0.4) &           79.6 (0.7) &    79.3 \\
luo      &           68.1 (0.9) &           69.2 (4.9) &  69.4 (2.2) &  \textbf{71.7 (0.9)} &    69.6 \\
pcm      &           85.5 (0.6) &  \textbf{89.2 (0.3)} &  86.2 (1.5) &           88.0 (0.1) &    87.2 \\
swa      &           87.5 (0.6) &  \textbf{88.3 (0.2)} &  87.5 (0.6) &           86.0 (0.7) &    87.3 \\
wol      &           61.4 (1.4) &  \textbf{66.1 (1.6)} &  63.9 (0.8) &           63.4 (0.9) &    63.7 \\
yor      &           79.3 (0.6) &  \textbf{80.9 (1.0)} &  76.5 (1.1) &           78.7 (0.7) &    78.8 \\
\midrule
Average  &                 79.3 &                 81.8 &        79.2 &                 73.0 &    78.3 \\
\bottomrule
\end{tabular}

    \end{adjustbox}
\end{table}

\subsection{How Corruption Affects Performance}
\begin{figure}[h]
    \centering
    \includegraphics[width=1\linewidth]{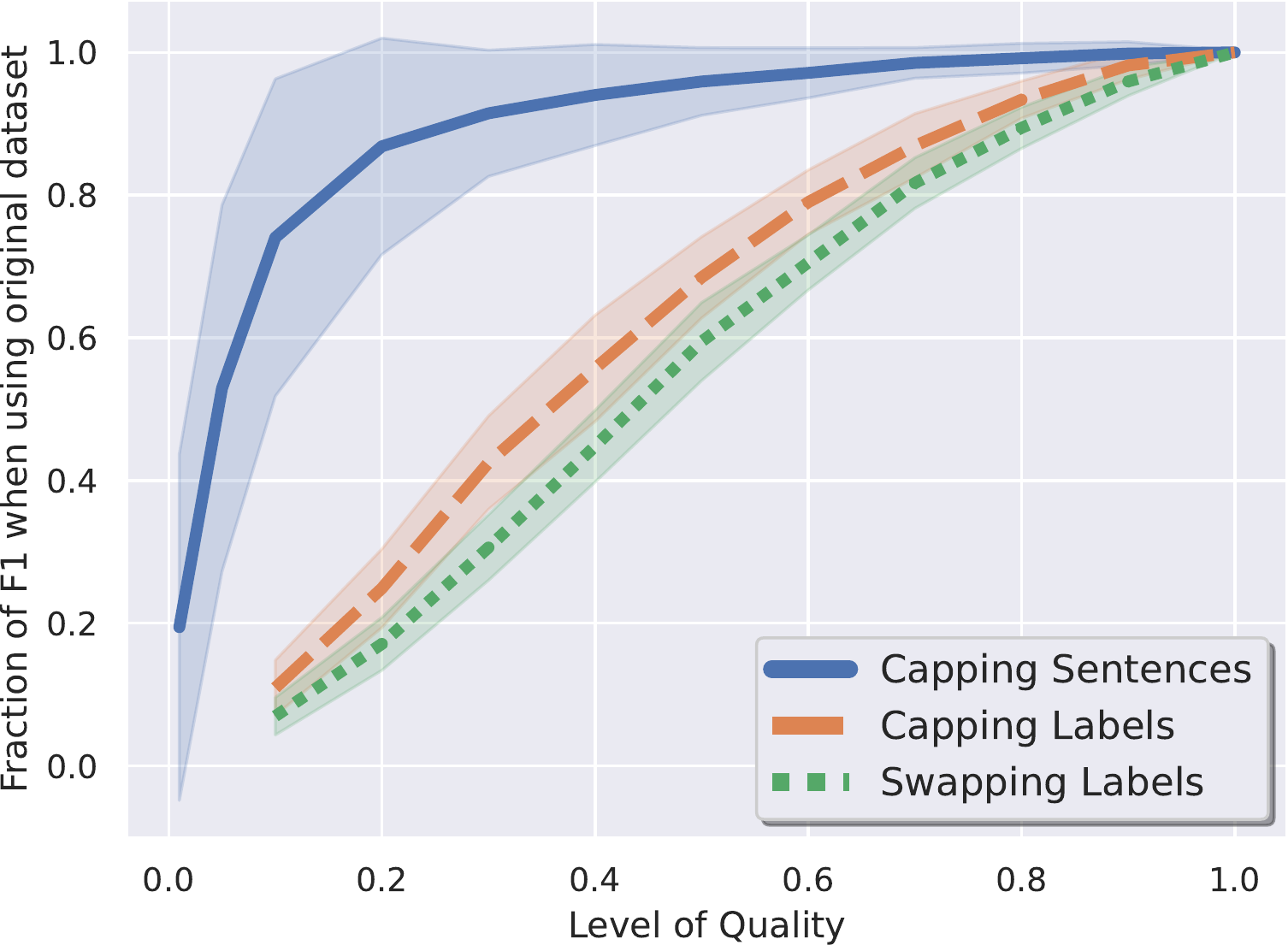}
    \caption{Comparing the result of (blue) deleting sentences, (orange) deleting labels and (green) swapping labels. The X-axis represents the level of quality: $1.0$ is the original dataset, whereas $0.1$ means that we keep $10\%$ of the sentences or $10\%$ of the labels and corrupt/delete the other 90\%. The mean here is shown, with the standard deviation across seeds, languages and models shaded.} 
    \label{fig:compare_global}
\end{figure}

Here we compare the three different corruption strategies: (1) deleting a certain fraction of sentences, (2) deleting (i.e., setting to \myO) a certain fraction of labels, and (3) swapping (i.e., replacing with another, but incorrect entity) a certain fraction of labels. The results are shown in \autoref{fig:compare_global}. The first conclusion we can draw from this is that the number of sentences is far less important than the quality of annotations. In particular, when deleting 90\% of the sentences (leaving us with about 10\% of the labels of the original dataset), we can still recover around 75\% of the performance of training on the entire dataset. When we set 90\% of the labels to \myO, however, the performance is much worse, just above 10\% of training on the original dataset. Thus, even though the number of labels is roughly equal for each case, having incorrectly labelled data affects the models much more than having fewer sentences that are completely labelled.\footnote{\label{note1}We do verify that when keeping only $X\%$ of the sentences, the fraction of labels we are left with is very close to $X\%$ compared to the original dataset, confirming that labels are mostly distributed uniformly throughout the sentences.}
When we swap labels with incorrect entities, the models perform similarly, but slightly worse compared to replacing these labels with \myO. This suggests that when an annotator is uncertain, it is better to leave an entity out compared to labelling it incorrectly.

Overall, this experiment shows that the fraction of correct annotations is an important factor in NER.
This means that, for every amount of labels, having fewer, but completely labelled sentences is significantly better than having the same number of sentences, but with incomplete or incorrect labels.

\subsection{Performance Across Models and Languages}
\begin{figure*}
        \centering
        \includegraphics[width=0.9\linewidth]{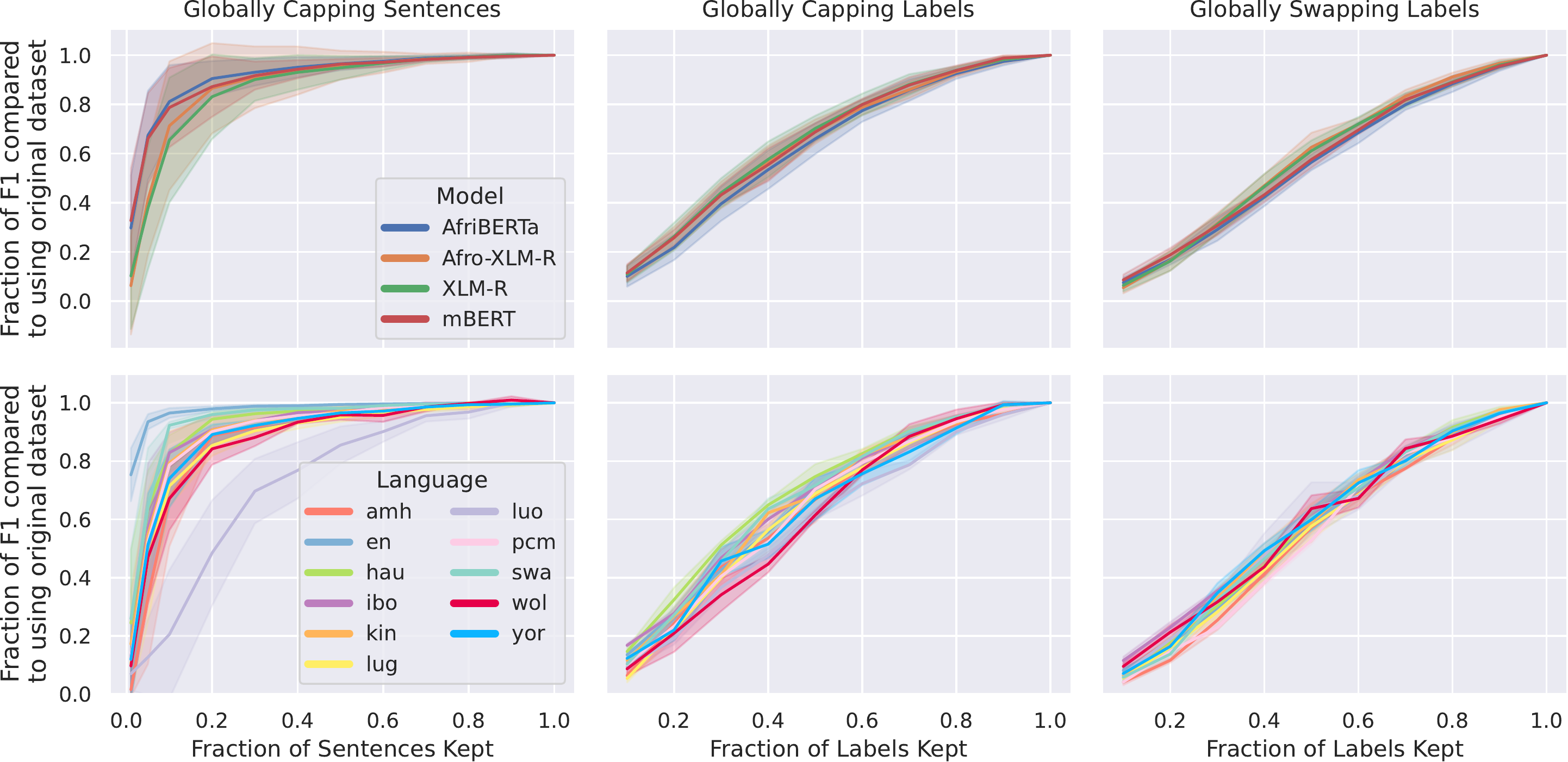}
    \caption{Showcasing the effect of each corruption when isolating each (top) pre-trained model or (bottom) language. We plot the mean and standard deviation over random seeds and the other feature (the language in  the top row and the model in the bottom row).}
    \label{fig:influence:all}
\end{figure*}

Having demonstrated that, on average, the quality of data is much more important than the quantity of data, we now investigate how these results differ across pre-trained models and languages. In the top row of \autoref{fig:influence:all} we plot the results of each corruption strategy, showing the performance of each model separately. 

Overall, all models perform similarly, with \afriberta and mBERT being slightly more robust to being trained on small amounts of data than the larger XLM-R models. We study the effect of language on our results in \autoref{fig:influence:all} (bottom), averaging over the random seeds and pre-trained models. We find that, on average, each language performs similarly. The only exception here is when we reduce the number of sentences the models were trained on, English performs better than average, whereas Luo performs worse. One reason for these observations is that the CONLL dataset is significantly larger than the other languages' datasets, making it perform well when given only a small fraction of the data, as this still corresponds to a large absolute number of sentences. Similarly, Luo has the smallest dataset out of all eleven languages, making it more susceptible to having even less data. This also suggests that having fewer, higher-quality sentences is preferred, but that having too few sentences can drastically reduce performance.

\subsection{The Tradeoff Between Quantity and Quality} 

\begin{figure*}[h]
    \centering
        \includegraphics[width=0.9\linewidth]{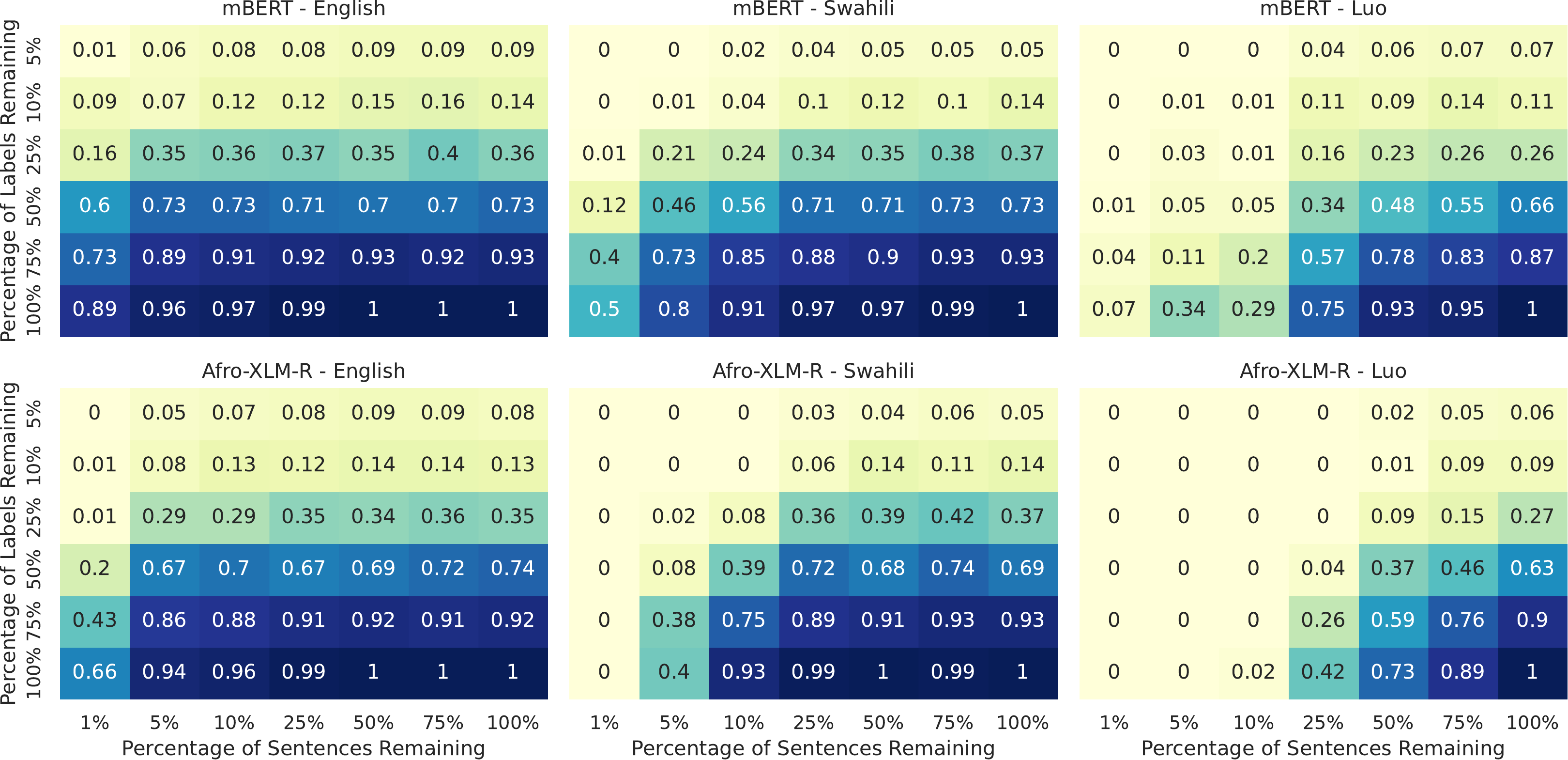}
    \caption{Showing the effect of varying both the number of sentences and the fraction of deleted labels on (top) mBERT and (bottom) \afroxlmr for \textit{en, swa} and \textit{luo}. The x-axis shows the percentage of sentences remaining whereas the y-axis lists the percentage of labels remaining. For example, at (50\%, 50\%), we first remove half of the sentences and then delete half of the labels in the remaining sentences. Each cell contains the fraction of F1 obtained when training on this data compared to training on the original dataset, averaged over 3 seeds.}
    \label{fig:lineplots:all}
\end{figure*}

The results in the previous sections indicate that prioritising data quality and the correctness of annotations is important, more so than the number of sentences we train on. We investigate this further by looking at the relationship between quantity and quality, combining the removal of sentences with the deletion of labels. Since we have shown that our results are mostly consistent across languages and models, we consider only \afroxlmr and mBERT as well as three languages, English, Swahili and Luo. 
We choose these two models due to their differences in number of parameters and pre-training languages. Most languages exhibit roughly similar performance, but English and Luo had slightly different behaviour when deleting sentences, due to their different dataset sizes. Swahili is used as a baseline as it had roughly average behaviour. The results are shown in \autoref{fig:lineplots:all} and they confirm that deleting labels is more detrimental to performance than removing a similar percentage of sentences. For example, for \afroxlmr and Swahili, keeping 25\% of the sentences but not removing any labels results in close to optimal performance, at 99\%. Having 25\% of the labels but keeping all of the sentences gives much worse performance, at 37\%, even though the overall number of correctly labelled entities is similar.\footnoteref{note1} Having 50\% of the sentences and 50\% of the labels, the results are in-between, just below 70\%. 
However, as can be seen when the fraction of remaining sentences becomes too small (e.g. having 10\% or fewer sentences for Luo), the models' performance suffers drastically, regardless of quality.
Thus, while having correctly annotated labels is a priority, if we do not have enough data, even perfect-quality annotations will result in poor performance.\footnote{mBERT shows a similar trend to \afroxlmr but, as before, it is slightly more robust to being fine-tuned on small amounts of data.}

\section{Discussion \& Future Work}

Our work follows a recent trend of questioning whether more data is always better in NLP, even at the cost of quality~\citep{alabi2019Massive,abdulmumin2022Separating}. In contrast to many of these works, we systematically and quantifiably investigate the effects of reducing the quantity and quality of data.
Firstly, our findings demonstrate that the quantity of data used for training does not greatly impact final performance, as we were able to achieve around 80\% performance with just 10-20\% of the original dataset's sentences. However, removing entity labels, which simulates a reduction in annotation quality, had a significant impact on performance. 
This supports prior findings that we do not need a large amount of data to perform well when leveraging pre-trained language models~\citep{adelani2022Thousand}. 
We do note, however, that when the number of sentences falls below a certain threshold (roughly between 200 and 400 sentences), performance drops significantly, indicating that we do need a minimum amount of data to perform well.
Furthermore, while quality has anecdotally been shown to be more important than quantity of data~\citep{alabi2019Massive,abdulmumin2022Separating}, here we quantify this effect in NER. Our results imply that when we have the budget to label $N$ entities, using fewer fully-labelled sentences is better than using more sentences that are only partially or incorrectly labelled.
Our results also suggest that even modest data-collection and annotation efforts should be able to result in datasets that are large enough to obtain decent performance. Quality, however, is of great importance and should be prioritised in the data creation process.

The second overarching observation we can make is that, in most cases, all models exhibit roughly equal behaviour, in terms of the dropoff in performance, as the level of corruption increases. This ranges from the African language-centric models (\afriberta and \afroxlmr) to the predominantly high-resourced models (\xlmr and mBERT). This suggests that the behaviour we see here is quite general, as opposed to being specific to just a particular model. The variation across the eleven languages is also remarkably low, again highlighting the consistency of our results. This is notable as we considered languages that were included in the pre-trained models' training data as well as some that were not.

There are numerous avenues for future work. One option would be to expand our work into other NLP tasks such as machine translation or question answering, to determine whether the same trend holds. Developing new corruption strategies that are applicable to other tasks and cover other aspects of quality would also be necessary to achieve this.
Investigating the effect of the corruptions on data characteristics, such as the prevalence of rare tokens, would also be valuable.
Furthermore, while we do not directly consider methods to address the problems introduced by corrupted data, our work motivates research into developing better methods for dealing with corrupted datasets, mitigating some of the negative effects of training on low-quality data. One promising option would be to use active learning, to choose which sentences should be labelled, or verified by a human annotator~\citep{gal2017Deep}.

Finally, we randomly remove sentences and find that this can still result in high performance. Recent work, however, has demonstrated that carefully choosing the data points to train on can result in significantly more data efficiency, requiring vastly less data while achieving comparable performance~\citep{mindermann2022Prioritized,sorscher2022Beyond}. This line of work is promising and has great potential in NLP for low-resourced languages.

\section{Conclusion}
In this paper, we present a systematic analysis of the impact of data quality and quantity on the performance of pre-trained models in a named entity recognition task for low-resourced languages. By designing multiple corruption strategies and fine-tuning models on datasets with varying degrees of corruption, we are able to provide useful insights into the relationship between data quality and model performance. 

Our results, which are consistent across pre-trained models and languages, demonstrate that pre-trained models can perform effectively with minimal data and that missing or incorrect annotations have a much greater negative impact than having fewer fully-labelled sentences. The findings of this study have the potential to inform future NER dataset creation efforts and aid NLP practitioners in selecting appropriate datasets for fine-tuning. 


\bibliographystyle{acl_natbib}
\bibliography{ijcai22,bib2,bib3,bib4}

\end{document}